\documentclass{article}

\usepackage[preprint]{nips_2018}

\usepackage[utf8]{inputenc} 
\usepackage[T1]{fontenc}    
\usepackage{hyperref}       
\usepackage{url}            
\usepackage{booktabs}       
\usepackage{amsfonts}       
\usepackage{nicefrac}       
\usepackage{microtype}      

\usepackage{amsmath}
\usepackage{graphicx}
\usepackage[colorinlistoftodos]{todonotes}
\usepackage{bm}
\usepackage{subfigure}
\usepackage{caption}

\newcommand{\Add}{\text{Add}}
\newcommand{\Mul}{\text{Mul}}
\newcommand{\AddMul}{\text{MulMix}}

\usepackage[
  separate-uncertainty = true,
  multi-part-units = repeat
]{siunitx}

\title{Learn to Combine Modalities in Multimodal Deep Learning}

\author{
  Kuan Liu$^{\dagger,\ddagger}$\thanks{Majority of the work in
this paper was carried out while the author was affiliated with Snap Research. Snap. Inc.} \\
  \texttt{kuanl@usc.edu} \\
  \And
  Yanen Li$^{\dagger}$\\
  \texttt{yanen.li@snap.com} \\
  \And
  Ning Xu$^{\dagger}$\\
  \texttt{ning.xu@snap.com}
  \AND
  Prem Natarajan$^{\ddagger}$ \\
  \texttt{pnataraj@isi.edu} \\
  $^{\dagger}$ Snap Research, Snap. Inc., Venice, CA \\
  $^{\ddagger}$ Computer Science Department \& Information Sciences Institute, 
  University of Southern California\\
}

\begin{document}

\maketitle

\begin{abstract}
  Combining complementary information from multiple modalities is intuitively appealing for improving the performance of learning-based approaches.  However, it is challenging to fully leverage different modalities due to practical challenges such as varying levels of noise and conflicts between modalities. Existing methods do not adopt a joint approach to capturing synergies between the modalities while simultaneously filtering noise and resolving conflicts on a per sample basis.  In this work we propose a novel deep neural network based technique that multiplicatively combines information from different source modalities. Thus the model training process automatically focuses on information from more reliable modalities while reducing emphasis on the less reliable modalities. Furthermore, we propose an extension that multiplicatively combines not only the single-source modalities, but a set of mixtured source modalities to better capture cross-modal signal correlations. We demonstrate the effectiveness of our proposed technique by presenting empirical results on three multimodal classification tasks from different domains.  The results show consistent accuracy improvements on all three tasks.
\end{abstract}

\section{Introduction}
Signals from different modalities often carry complementary information about different aspects of an object, event, or activity of interest. Therefore, learning-based methods that combine information from multiple modalities are, in principle, capable of more robust inference.  For example, a  person's visual appearance and the type of language he uses both carry information about his age. In the context of user profiling in a social network, it helps to predict users' gender and age by modeling both users' profile pictures and their posts. A natural generalization of this idea is to aggregate signals from all available modalities and build learning models on top of the aggregated information, ideally allowing the learning technique to figure out the relative emphases to be placed on different modalities for a specific task. This idea is ubiquitous in existing multimodal techniques, including early and late fusion~\cite{snoek2005early,gunes2005affect}, hybrid fusion~\cite{atrey2010multimodal}, model ensemble~\cite{dietterich2000ensemble}, and more recently---joint training methods based on deep neural networks~\cite{ngiam2011multimodal,wollmer2010context,neverova2016moddrop}. In these methods, features (or intermediate features) are put together and are jointly modeled to make a decision. We call them \textit{additive} approaches due to the type of aggregation operation. Intuitively, they are able to gather useful information and make predictions collectively.

However, it is practically challenging to learn to combine different modalities. Given multiple input modalities, artifacts such as noise may be a function of the sample as well as the modality; for example, a clear, high-resolution photo may lead to a more confident estimation of age than a lower quality photo. Also, either signal noise or classifier vulnerabilities may result in decisions that lead to conflicts between modalities. For instance, in the example of user profiling, some users' gender and age can be accurately predicted by a clear profile photo, while others with a noisy or otherwise unhelpful (e.g., cartoon) profile photo may instead have the most relevant information encoded in their social network engagement---such as posts and friend interactions, etc. In such a scenario, we refer to the affected modality, in this case the image modality, as a \textit{weak modality}. We emphasize that this weakness can be sample-dependent, and is thus not easily controlled with some global bias parameters. An ideal algorithm should be robust to the noise from those weak modalities and pick out the relevant information from the strong modalities on a \textit{per sample basis,} while at the same time capturing the possible complementariness among modalities.

We would like to point out that the existing \textit{additive} approaches do not fully address the challenges mentioned earlier. Their basic assumptions are 1) every modality is always potentially useful and should be aggregated, and 2) the models (e.g., a neural network) on top of aggregated features can be trained well enough to recover the complex function mapping to a desired output. While in theory the second assumption should hold, i.e., the learned models should be able to determine the quality of each modality per sample if given a sufficiently large amount of data. They are, in practice, difficult to train and regularize due to the finiteness of available data.

In this work, we propose a new \textit{multiplicative} multimodal method which explicitly models the fact that on any particular sample not all modalities may be equally useful. The method first makes decisions on each modality independently. Then the multimodal combination is done in a differentiable and \textit{multiplicative} fashion. This multiplicative combination suppresses the cost associated with the weak modalities and encourages the discovery of truly important patterns from informative modalities. In this way, on 
a particular sample, inferences from weak modalities gets suppressed in the final output. And even more importantly perhaps, \textit{they are not forced to generate a correct prediction (from noise!) during training}. This accommodation of weak modalities helps to reduce model overfitting, especially to noise. As a consequence, the method effectively achieves an automatic selection of the more reliable modalities and ignores the less reliable ones. The proposed method is also end-to-end and enables jointly training model components on different modalities.

Furthermore, we extend the \textit{multiplicative} method with the ideas of \textit{additive} approaches to increase model capacity. The motivation is that certain unknown \textit{mixtures} of modalities may be more useful than a good single modality. The new method first creates different mixtures of modalities as candidates, and they make decisions independently. Then the multiplicative combination automatically selects more appropriate candidates. In this way, the selection operates on "modality mixtures" instead of just a single modality. This mixture-based approach enables structured discovery of the possible correlations and complementariness across modalities and increases the model capacity in the first step. A similar selection process applied in the second step ignores irrelevant and/or redundant modality mixtures. This again helps control model complexity and avoid excessive overfitting.

We validate our approach on classification tasks in three datasets from different domains: image recognition, physical process classification, and user profiling. Each task provides more than one modality as input. Our methods consistently outperform the existing, state-of-the-art multimodal methods.

In summary, the key contributions of this paper are as follows:

\begin{itemize}
\item The multimodal classification problem is considered with a focus on addressing the challenge of weak modalities.
\item A novel deep learning combination method that automatically selects strong modalities per sample and ignores weak modalities is proposed and experimentally evaluated. The method works with different neural network architectures and is jointly trained in an end-to-end fashion.
\item A novel method to automatically select mixtures of modalities is presented and evaluated. This method increases model capacity to capture possible correlations and complementariness across modalities. 
\item Experimental evaluations on three real-world datasets from different domains show that the new methods consistently outperform existing multimodal methods.
\end{itemize}

\begin{figure*}
\centering
\includegraphics[width=1.0\textwidth]{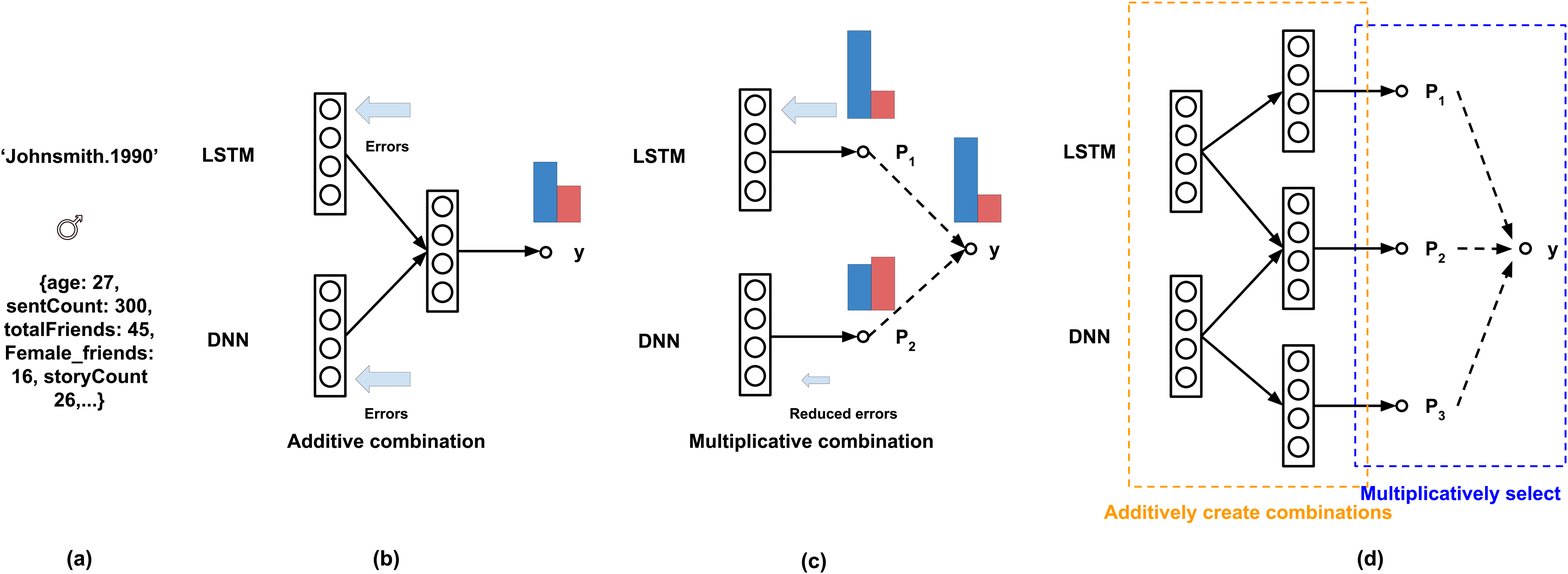}
\caption{\label{fig:illus} Illustration of different deep neural network based multimodal  methods. (a) A gender prediction example with text (a fake userid) and dense feature (fake profile information) modality inputs.  (b) Additive combination methods train neural networks on top of aggregate signals from different modalities; Equal errors are back-propagated to different modality models. (c) Multiplicative combination selects a decision from a more reliable modality; Errors back to the weaker modality are suppressed. (d) Multiplicative modality mixture combination first additively creates mixture candidates and then selects useful modality mixtures with multiplicative combination procedure.}
\end{figure*}

\section{Background}

To set the context for our work, we now describe two existing types of popular multimodal approaches 
in this section. We begin first with notations and then describe traditional approaches, followed by existing deep learning approaches. 

\paragraph{Notations}
We use $M$ to indicate the number of modalities available in total. We denote each input modality/signal as a dense vector $\bm{v}_m \in \mathbb{R}^{d_m}, \forall m=1,2,..,M$. For example, given $M=3$ modalities in the user profiling task, $\bm{v}_1$ is the profile image represented as a vector, $\bm{v}_2$ is the posted text representation, and $\bm{v}_3$ encodes the friend network information. We consider a K-way classification setting where $y$ denotes the labels. $p_m^k$ denotes the prediction probability of the $k$th class from the $m$th modality, and $p^k$ denotes the model's final prediction probability of the $k$th class. Throughout the paper, superscripts are used with indices for classes and subscripts used for modalities.

\subsection{Traditional Approaches}
\noindent\textbf{Early Fusion}
Early fusion methods create a joint representation of input features from multiple modalities. Next, a single model is trained to learn the correlation and interactions between low level features of each modality. We denote the single model as $h$. The final prediction can be written as 
\begin{equation}
\label{eq:early}
p = h([\bm{v}_1, .., \bm{v}_m]),
\end{equation}
where we use concatenation here as a commonly seen example of jointly representing modality features.

Early fusion could be seen as an initial attempt to perform multimodal learning. The training pipeline is simple as only one model is involved. It usually requires the features from different modalities to be highly engineered and preprocessed so that they align well or are similar in their semantics. Furthermore, it uses one single model to make predictions, which assumes that the model is well suited for all the modalities.

\noindent\textbf{Late Fusion}
Late fusion uses unimodal decision values and fuses them with a fusion mechanism $F$ (such as averaging~\cite{shutova2016black}, voting~\cite{morvant2014majority}, or a learned model \cite{glodek2011multiple,ramirez2011modeling}.) Suppose model $h_i$ is used on modality $i$ ($i=1,..,M$,) the final prediction is


\begin{equation}
\label{eq:late}
p = F(h_1(\bm{v}_1), ..., h_m(\bm{v}_m))
\end{equation}

Late fusion allows the use of different models on different modalities, thus allowing more flexibility. It is easier to handle a missing modality as the predictions are made separately. However, because late fusion operates on inferences and not the raw inputs, it is not effective at modeling signal-level interactions between modalities.

\subsection{Multimodal Deep Learning}
Due to the superior performance and computationally tractable representation capability (in vector spaces) in multiple domains such as visual, audio, and text, deep neural networks have gained tremendous popularity in multimodal learning tasks \cite{ngiam2011multimodal,ouyang2014multi,wang2015deep}. Typically, domain-specific neural networks are used on different modalities to generate their representations and the individual representations are merged or aggregated. Finally, the prediction is made on top of aggregated representation usually with another a neural network to capture the interactions between modalities and learn complex function mapping between input and output. Addition (or average) and concatenation are two common aggregation methods, i.e.,

\begin{equation}
\label{eq:add1}
\bm{u} = \sum_m f_m(\bm{v}_m)
\end{equation}
or
\begin{equation}
\label{eq:add2}
\bm{u} = [f_1(\bm{v}_1), .., f_1(\bm{v}_m)]
\end{equation}
where $f$ is considered a domain specific neural network and $f_m: \mathbb{R}^{d_m}\rightarrow \mathbb{R}^{d} (m=1,..,M)$. Given the combined vector output $u \in \mathbb{R}^d$ (or $\mathbb{R}^{\sum d_m}$), another network $g$ computes the final output.

\begin{equation}
\label{eq:add_output}
p = g(\bm{u}) \quad where \quad g: \mathbb{R}^{d}\rightarrow \mathbb{R}^{K}
\end{equation}

The network structure is illustrated in Figure \ref{fig:illus}(b) . The arrows are function mappings or computing operations. The dotted boxes are representations of single and combined modality features. We call them \textbf{additive combinations} because their critical step is to add modality hidden vectors (although often in a nonlinear way).

In Section 5, we present related work in areas such as learning joint multimodal representations using a shared semantic space. Those approaches are not directly applicable to our task where we are aim to predict latent attributes, not merely the observed identities of the sample.


\section{A multiplicative combination layer}
\label{sec:mul}

The \textit{additive} approaches discussed above make no assumptions regarding the reliability of different modality inputs. As such, their performance critically relies on the single network $g$ to figure out the relative emphases to be placed on different modalities. From a modeling perspective, the aim is to recover the function mapping between the combined representation $u$ and the desired outputs. This function can be complex in real scenarios. For instance, when the signals are similar or complementary to each other, $g$ is supposed to merge them to make a strengthened decision; when signals conflict with each other, $g$ should filter out the unreliable ones and make a decision based primarily on more reliable modalities. While in theory $g$---often parameterized as a deep neural network---has the capability to recover an arbitrary function given a sufficiently large amount of data (essentially, unlimited), it can be, in practice, very difficult to train and regularize given data constraints in real applications. As a result, model performance degrades significantly.

Our aim is to design a more (statistically) efficient method by explicitly assuming that \textit{some modalities are not as informative as others} on a particular sample. As a result, they should not be fed into a single network for training. Intuitively, it is easier to train a model on the input of a good modality rather than a mix of good ones and bad ones. Here we differentiate modalities to be \textit{informative modalities} (good) and \textit{weak modalities} (bad). Note that the labels informative and noisy are applied in respect to each particular sample.

To begin, let every modality make its own independent decision with its modal-specific model (e.g., $p_i = g_i(\bm{v}_i)$.) Their decisions are combined by taking an average. Specifically, we have the following objective function,

\begin{equation}
\label{eq:ce}
L_{ce} = \ell^y_{ce},  \qquad \ell^y_{ce} = - \sum_{i=1}^M \log p_i ^y
\end{equation}
where $y$ denotes the true class index, and we call $\ell^y$ a \textit{class loss} as it is part of the loss function associated with a particular class. In the testing stage, the model predicts the class with the smallest class loss, i.e.,

\begin{equation}
\label{eq:ce_test}
\hat{y} = arg\min_y \ell^y_{ce}.
\end{equation}

This relatively standard approach allows us to train one model per modality. However, when weak modalities exist, the objective (\ref{eq:ce}) would significantly increase. By minimizing (\ref{eq:ce}), it forces every model based on its modality to perform well on the training data. This could lead to severe overfitting as the noisy modality simply does not contain the information required to make a correct prediction, but the loss function penalizes it heavily for incorrect predictions. 

\subsection{Combine in a multiplicative way}

To mitigate against the problem of overfitting, we propose a mechanism to suppress the penalty incurred on noisy signals from certain modalities. A cost on a modality is down-weighted when there exists other good modalites for this example. Specifically, a modality is good (or bad) when it assigns a high (or low) probability to the correct class. A higher probability indicates more informative signals and stronger confidence. With that in mind, we design a down-weighting factor as follows,

\begin{equation}
\label{eq:scale}
q_i = [\prod_{j\neq i} (1-p_j) ]^{\beta/(M-1)}
\end{equation}
where we omit the class index superscripts on $p$ and $q$ for brevity; $\beta$ is a hyper parameter to control the strength of down-weighting and are chosen by cross-validation. Then the new training criterion becomes 
\begin{equation}
\label{eq:mul}
L_{mul} =  \ell^y_{mul}, \qquad  \ell^y_{mul} = -  \sum_{i=1}^M q^y_i  \log p^y_i.
\end{equation}


The scaling factor $[\prod_{j\neq i} (1-p_j) ]^{\beta/(M-1)}$ represents the average prediction quality of the remaining modalities. This term is close to 0 when some $p_j$ are close to 1. When those modalities ($j\neq i$) have confident predictions on the correct class, the term would have a small value, thus suppressing the cost on the current modality ($p_i$). Intuitively, when other modalities are already good, the current modality ($p_i$) \textit{does not have to be equally good}. This down-weighting reduces the training requirement on all modalities and reduces overfitting. \cite{jin2016collaborative} uses this term to ensemble different layers of a convolution network in image recognition task. We introduce important hyper-parameter $\beta$ to control the strength of these factors. Larger values give a stronger suppressing effect and vice versa. During the testing, we follow a similar criterion in (\ref{eq:ce_test}) (replacing $\ell_{ce}$ with $\ell_{mul}$.)

We call this strategy a \textbf{multiplicative combination} due to the use of multiplicative operations in (\ref{eq:scale}). 
During the training, the process always tries to select some modalities that give the correct prediction and tolerate mistakes made by other modalities. This tolerance encourages each modality to work best in its own areas instead of on all examples.

We emphasize that $\beta$ implements a trade off between ensemble and non-smoothed multiplicative combination. When $\beta$=0, we have $q=1.0$ and predictions from different modalities are averaged; when $\beta$=1, there is no smoothing at all on $(1-p_j)$ terms so that a good modality will strongly down weight losses from other modalities.

The proposed combination can be implemented as the last layer of a combination neural network as it is differentiable. Errors in (\ref{eq:mul}) can be back-propagated to different components of the model such that the model can be trained jointly.

\subsection{Boosted multiplicative training}
\label{sec:boosted}

Despite it providing a mechanism to selectively combine good and bad modalities, the  multiplicative layer as configured above has some limitations. Specifically, there is \textit{a mismatch between minimizing the objective function and maximizing the desired accuracy}. To illustrate that, we take one step back and look at the standard cross entropy objective function (\ref{eq:ce}) (then $M = 1$). We have $\exp(-\ell^1) + \exp(-\ell^2) = p^1 + p^2 = 1$ when $Y = 2$. Let's call it \textit{normalized}. It makes intuitive sense to minimize only $\ell^y$ in the training phase so that we have $\ell^y < \ell^{y'} (y'\neq y)$---thus maximizing the accuracy.

However, if we look at (\ref{eq:mul}), the same normalization does not apply any more due to the complication of multiple modalities ($M>1$) and the introduction of the down-weighting factors $q_i$. Therefore, it does not guarantee that minimizing $\ell^1$ leads to driving $\ell^1<\ell^2$ or vice versa. There are two important consequences of this mismatch. First, the method may stop minimizing the class losses on the correct classes when it is still incorrect. Second, it may work on reducing the class losses that already have correct predictions.

\noindent\textbf{A tempting naive approach} When addressing the issue, a temptive approach is to normalize the class losses similar to normalizing a probability vector. A deeper consideration reveals the pitfall inherent in that temptation---normalizing class losses does not make sense because the class losses in an objective function are error surrogates which usually serve as the upper bound of the training errors. While it makes sense to minimize the surrogates on the correct classes, but it is pointless, perhaps counterproductive, to maximize the losses on the wrong classes. What regular normalization techniques do is to maximize the gap between losses in the correct and wrong classes---effectively minimizing the correct ones and maximizing the wrong ones. Experimental results validate the analysis presented above. 

\noindent\textbf{Boosting extension} We propose a modification of the objective function in (\ref{eq:mul}) to address the issue. Rather than always placing a loss on the correct class, we place a penalty only when the class loss values are not the smallest among all the classes. This creates a connection to the prediction mechanism in (\ref{eq:ce_test}). If the prediction is correct, there is no need to further reduce the class loss on that instance; if the prediction is wrong, the class loss should be reduced even if the loss value is already relatively small. To increase the robustness, we add a margin formulation where the loss on the correct class should be smaller by a margin. Thus, the objective we use is as follows, 

\begin{equation}
\label{eq:mulp}
L = \ell^y (1 - \prod_{\forall y' \neq y}1(\ell_{mul}^y + \delta  < \ell_{mul}^{y'}) ),
\end{equation}
where the bracket part in the right-hand side of (\ref{eq:mulp}) computes whether the loss associated with the correct class is the smallest (by a margin). The margin $\delta$ is chosen in experiment by cross validation.

The new objective function only aims to minimize the class losses which still need improvement. For those examples that already have correct classification, the loss is counted as zero. Therefore, the objective only adjusts the losses that lead to wrong prediction. It makes model training and desired prediction accuracy better aligned.

\noindent\textbf{Boosting connection} There is a clear connection between the new objective (\ref{eq:mulp}) and boosting ideas if we consider the examples where (\ref{eq:ce_test}) makes wrong predictions as hard examples and others as easy examples. The objective (\ref{eq:mulp}) looks at only the hard examples and directs efforts to improve the losses. The hard examples change during the training process, and the algorithm adapts to that. Therefore, we call the new training criterion boosted multiplicative training.

\section{Select modality mixtures}
\label{sec:modalmix}
The \textit{multiplicative} combination layer explicitly assumes modalities are noisy and automatically select good modalities. One limitation is that the models $g_i$ ($i=1,..,M$) are trained primarily based on a single modality (although they do receive back-propagated errors from the other modalities through joint training). This prevents the method from fully capturing synergies across modalities. In a twitter example, a user's follower network and followee network are two modalities that are different but closely related. They jointly contribute to predictions concerning the user's interests, etc. The multiplicative combination in Section \ref{sec:mul} would not be ideal in capturing such correlations. On the other hand, \textit{additive} methods are able to capture model correlation more easily by design (although they do not explicitly handle modality noise and conflicts).

\subsection{Modality mixture candidates}
Given the complementary qualities of the additive and multiplicative approaches, it is desirable to harness the advantages of both. To achieve that goal, we propose a new method. At a high level, we want our methods to first have the capability to capture all possible interactions between different modalities and then to filter out noises and pick useful signals.

In order to be able to model interactions of different modalities, we first create \textbf{different mixtures of modalities}. Particularly, we enumerate all possible mixtures from the power set of the set of modality features. On each mixture, we apply the \textit{additive} operation to extract higher-level feature representations as follows,

\begin{equation}
\label{eq:comb_m}
\bm{u}_c = \sum_{k \in M_c; M_c \subset \{1,2,..,M\}} f(\bm{v}_k)
\end{equation}
where $M_c$ contains one or more modalities. Thus we have $u_c$ as the representation of the mixture of modalities in set $M_c$. It gathers signals from all the modalities in $M_c$. Since there are $2^M - 1$ different non-empty $M_c$, there are $2^M - 1$ $u_c$, and each $u_c$ looks into the mix of a different modality mixture. We call each $u_c$ a \textbf{mixture candidate} as we believe not every mixture is equally useful; some mixtures may be very helpful to model training while others could even be harmful.

Given the generated candidates, we make predictions based on each of them independently. Concretely, as in the \textit{additive} approach, a neural network is used to make prediction $p_c$ as follows,
\begin{equation}
\label{eq:comp_m}
p_c = g_c(\bm{u}_c),
\end{equation}.
where $p_c$ is the prediction result from an individual mixture. Different $p_c$ may not agree with each other. It remains to have a mechanism to select which one to believe or how to combine $p_c$.

\subsection{Mixture selections}

Among the combination candidates generated above, it is not clear which mixtures are strong and which are weak due to the way of enumerating proposals. One simple way is to average predictions from all candidates. However, it loses the ability to discriminate between different modalities and again takes modalities as equally useful. From the modeling perspective, it is similar to simply doing additive approaches to modalities in the beginning. Our goal is to automatically select strong candidates and ignore weak ones. 

To achieve that, we apply the \textit{multiplicative} combination layer (\ref{eq:mul}) in Section \ref{sec:mul} to the selection of mixture candidates in (\ref{eq:comp_m}), i.e.,

\begin{equation}
\label{eq:addmul}
\ell^y = -  \sum_{c=1}^{|M_c|} q^y_c  \log p^y_c
\end{equation}
where $q_c$ is defined similarly. 
Equation (\ref{eq:addmul}) follows (\ref{eq:mul}) except that each model here is based on a mixture candidate instead of a single modality. 

With (\ref{eq:mulp}) (\ref{eq:comb_m}) (\ref{eq:comp_m}) (\ref{eq:addmul}) our method pipeline can be illustrated in Fig. \ref{fig:illus}(d). It first additively creates modality mixture candidates. Such candidates can be features from one single modality and can also be mixed features from multiple modalities. These candidates by design make it more straightforward to consider signal correlation and complementariness across modalities. However, it is unknown which candidate is good for an example. Some candidates can be redundant and noisy. The method then combines the prediction of different mixtures multiplicatively. The multiplicative layer enables candidate selection in an automatic way where strong candidates are picked while weak ones are ignored without dramatically increasing the entire objective function. As a whole, the model is able to pick the most useful modalities and modality mixtures with respect to our prediction task.

\section{Related Work}

\noindent\textbf{Multimodal learning} Traditional multimodal learning methods include early fusion (i.e., feature based), late fusion (i.e., decision based), and hybrid fusion \cite{atrey2010multimodal}. They also include model based fusion such as multiple kernel learning \cite{gonen2011multiple,bucak2014multiple,gehler2009feature}, graphical model based approaches\cite{nefian2002coupled,ghahramani1996factorial,gurban2008dynamic}, etc. 

Deep neural networks are very actively explored in multimodal fusion \cite{ngiam2011multimodal}. They have been used to fuse information for audio-visual emotion
classification~\cite{wollmer2010context}, gesture recognition \cite{neverova2016moddrop}, affect analysis \cite{kahou2016emonets}, and video description generation~\cite{jin2016video}. While the modalities used, architectures, and optimization techniques might differ, the general idea of fusing information in joint hidden layer of a neural network remains the same. 

\noindent\textbf{Multiplicative combination technique} Multiplicative combination is widely explored in machine learning methods. \cite{changpinyo2013similarity} uses an OR graphical model to combine similarity probabilities across different feature components. The probabilities of dissimilarity between pairs of objects go through multiplication to generate the final probability of be dissimilar, thus picking out the most optimistic component. \cite{jin2016collaborative} ensembles multiple layers of a convolution network with a down-weighting objective function which is a specialized instance of our (\ref{eq:scale}). Our objective is more general and flexible. Furthermore, we developed boosted training strategy and modality combination to address multimodal classification challenges. \cite{lin2017focal} develops focal loss to address class imbalance issue. In its single modality setting, it down-weights every class loss by one minus its own probability.

Attention techniques \cite{ba2014multiple,chan2016listen,mnih2014recurrent} are also treated as multiplicative methods to combine multiple modalities. Features from different modalities are dynamically weighted before mixed together. The multiplicative operation is performed at the feature level instead of decision level.

\noindent\textbf{Other multimodal tasks}. There are other multimodal tasks where the ultimate task is not classification. These include such various image captioning tasks. In~\cite{vinyals2015show}~a CNN image representation is decoded using an LSTM language model. In~\cite{jia2015guiding}, gLSTM~incorporates the image data together with sentence decoding at every time step fusing visual and sentence data in a joint representation. Joint multimodal representation learning is also used for visual and media question answering~\cite{gao2015you,malinowski2015ask,xu2016ask}, visual integrity assessment~\cite{jaiswal2017multimedia}, and personalized recommendation~\cite{hidasi2016parallel,liu2017batch}.

\section{Experiments}

We validate our methods on three datasets from different domains: image recognition, physical process classification, and user profiling. On these tasks, we are given more than one modality inputs and try to best use these modalities to achieve good generalization performance. Our code is publicly available\footnote{\url{https://github.com/skywaLKer518/MultiplicativeMultimodal}}.


\subsection{Setup}

\begin{table}
\caption{\label{tab:data}Datasets and modalities.}
\centering
\begin{tabular}{l|c} \hline
Datasets & Modalities \\\hline
CIFAR100 & features output from 3 ResNet units \\
HIGGS & low-level, high-level features  \\
gender & first name, userid, engagement \\ \hline
\end{tabular}
\end{table}

\subsubsection{CIFAR-100 image recognition}

The CIFAR-100 dataset~\cite{krizhevsky2009learning}~contains 50,000 and 10,000 color images with size of 32 $\times$ 32 from 100 classes for training and testing purposes, respectively. As observed by \cite{yang2015multi,hariharan2015hypercolumns}, different layers of a convolutional neural network (CNN) contain different signals of an image (different abstraction levels) that may be useful for classification on different examples. 
\cite{jin2016collaborative}~takes three layers of networks in networks (NINs)~\cite{lin2013network}~and demonstrates recognition accuracy improvements. In our experiments, the features from three different layers of a CNN are regarded as three different modalities. 

\noindent\textbf{Network architecture} We use Resnet~\cite{he2016deep} as the network in our experiments as it significantly outperforms NINs on this task and meanwhile Resnet also has the block structure so that it makes our choice of modality easier and more natural. We experimented with network architecture Resnet-32, Resnet-110. On both networks there are three residual units. We take the hidden states of the three units as modalities. We follow \cite{jin2016collaborative} and weight the losses of different layers by (0.3, 0.3, 1.0). Our implementations are based on~\cite{gomez17revnet}. 

\noindent\textbf{Methods} We experimented different methods: (1) Vanilla Resnet (``Base'')\cite{gomez17revnet} predicts the image class only based on the last layer output; i.e., there is only one modality.  (2) Resnet-Add (``Add'') concatenates the hidden nodes of three layers and builds fully connected neural networks (FCNNs) on top of the concatenated features. We tuned the network structure and found a two layer network with hidden 256 nodes gives the best result. (3) Resnet-Mul (``Mul'') multiplicatively combines predictions from the hidden nodes of three layers. (4) Resnet-MulMix (``MulMix'') uses multiplicative modality mixture combination on the three hidden layers. It uses default $\beta$ value $0.5$. (5) Resnet-MulMix* (``MulMix*'') is the same as MulMix except $\beta$ is tuned between 0 and 1.


\noindent\textbf{Training details} We strictly follow~\cite{gomez17revnet} to train Resnet and all other models. Specifically, we use SGD momentum with a fixed schedule learning rate \{0.1, 0.01, 0.001\} and terminate at 80000 iterations. We use 100 as batch size and choose weight decay 0.0002 on all network weights.

\subsubsection{HIGGS classification}

HIGGS \cite{baldi2014searching} is a binary classification problem to distinguish between a signal process which produces Higgs bosons and a background process which does not. The data has been produced using Monte Carlo simulations. We have two feature modalities---low level and high level features. Low level features are 21 features which are kinematic properties measured by the particle detectors in the accelerator. High level features are anothor 7 features that are functions of the first 21 features; they are derived by physicists to help discriminate between the two classes. Details of feature names are in the original paper~\cite{baldi2014searching}. We follow the setup in~\cite{baldi2014searching}~and use the last 500,000 examples as a test set.


\noindent\textbf{``HIGGS-small'' and ``HIGGS-full''}. To investigate the algorithm behaviors under different data scales, we also randomly down-sample 1/3 of the examples from the entire \textit{training} split. This creates another subset which we call "HIGGS-Small."

\noindent\textbf{Network architecture} We use feed-forward deep networks on each modality. We follow~\cite{baldi2014searching}~to use 300 hidden nodes in our networks and tried different number of layers. L2 weight decay is used with coefficient 0.00002. Dropout is not used as it hurts the performance in our experiments. Network weights are randomly initialized and SGD with 0.9 momentum is used during the training.

\noindent\textbf{Methods} We experimented with single modality prediction, late fusion of two modalities, and modality combination methods similar to what we describe above in CIFAR100.
%
%

\subsubsection{Gender classification}
\noindent\textbf{Gender} The dataset we use contains 7.5 million users from Snapchat app, with registered userid and user activity logs (e.g., story posts, friend networks, message sent, etc.) It also contains the inferred user first names produced by an internal tool. The task is to predict user genders. Users' inputs in Bitmoji app are used as the ground truth. We randomly shuffle the data and use 6 million samples as training, 500K as development, and 1 million for testing.

There are three modalities in this dataset: userid as a short text, inferred user first name as a letter string, and dense features extracted from user activities (e.g., the count of message sent, the number of male or female friends, the count of stories posted.)

\noindent\textbf{Gender-6 and Gender-22}
We experimented with two versions of the dataset. The versions differed in the richness of user activity features. The first one has 6 highly engineered features (gender-6) and the other has 22 features (gender-22).

\noindent\textbf{Network architecture}
We use FCNNs to model dense features. We tune the architecture and eventually use a 2 layer network with 2000 hidden nodes. We use (character based) Long Short-Term Memory Networks (LSTMs)~\cite{hochreiter1997long}~to model text string. The text string is fed into the networks one character by one character and the hidden representation of the last character is connected with FCNNs to predict the gender. We find the vanilla single layer LSTMs outperforms or matches other variants including (multi-layer, bidirectional~\cite{graves2013speech}, attention-based LSTMs). We believe it is due to the fact that we have sufficiently large amount of data. We also experimented character based Convolutional Neural Networks (char-CNN)~\cite{kim2014convolutional,zhang2015character}~and CNNs+LSTMs for text modeling and found LSTMs perform slightly better. 

Training details: our tuned LSTMs has 1 layer with hidden size 512. It uses ADAM~\cite{kingma2014adam}~to train with learning rate 0.0001 and learning rate decay 0.99. Gradients are truncated at 5.0. We stop model training when there is no improvement on the development set for consecutive $15$ evaluations.

\noindent\textbf{Methods} In addition to methods described in CIFAR100 and HIGGS, we also experimented with an attention based combination methods \cite{moon2018multimodal}. 


\begin{table}[]
\centering
\caption{Test error rates/AUC comparisons on CIFAR100, HIGGS, and gender tasks. \AddMul~uses default $\beta$ value 0.5. \AddMul*~tunes $\beta$ between 0 and 1. Experimental results are from 5 random runs. The best and second best results in each row are in bold and italic, respectively.}
\label{tab:acc}
\begin{tabular}{|c||c|c|c||c|c|c|}
\hline
    & Base                                                                & Fuse & Add   \cite{ngiam2011multimodal}                                                              & Mul            & Mulmix        & Mulmix*     \\ \hline
\multicolumn{7}{|c|}{cifar100, resnet-32}                                                                                                                                                              \\ \hline
Err & \begin{tabular}[c]{@{}c@{}}30.3  $\pm$ 0.2\\ 30.0 ~\cite{gomez17revnet} \end{tabular} & -    & 29.4 $\pm$ 0.4                                                       & 29.3 $\pm$ 0.2   & \textit{27.8 $\pm$ 0.3}  & \textbf{27.3 $\pm$ 0.4}  \\ \hline
\multicolumn{7}{|c|}{cifar100, resnet110}                                                                                                                                                              \\ \hline
Err & \begin{tabular}[c]{@{}c@{}}26.5 $\pm$ 0.3\\ 26.4 ~\cite{gomez17revnet}\end{tabular}  & -    & 27.2 $\pm$ 0.4                                                       & 25.3 $\pm$ 0.4   & \textit{25.1 $\pm$ 0.2}  & \textbf{24.7 $\pm$ 0.3}  \\ \hline
\multicolumn{7}{|c|}{higgs-small}                                                                                                                                                                      \\ \hline
Err & 23.3                                                                & 22.5 & 22.3 $\pm$ 0.1                                                       & 21.8 $\pm$ 0.1   & \textit{21.4 $\pm$ 0.1} & \textbf{21.2 $\pm$ 0.1}  \\ \hline
AUC & 84.8                                                                & 85.9 & 86.2 $\pm$ 0.1                                                       & 86.5 $\pm$ 0.1   & \textit{87.1 $\pm$ 0.1} & \textbf{87.2 $\pm$ 0.1}  \\ \hline
\multicolumn{7}{|c|}{higgs-full}                                                                                                                                                                       \\ \hline
Err & 21.7                                                                & 20.6 & 20.0 $\pm$ 0.1                                                       & 20.1 $\pm$ 0.1   & \textit{19.6 $\pm$ 0.2}  & \textbf{19.4 $\pm$ 0.1}  \\ \hline
AUC & 86.6                                                                & 88.0 & \begin{tabular}[c]{@{}c@{}}88.6 $\pm$ 0.1\\ 88.5 ~\cite{baldi2014searching} \end{tabular} & 88.3 $\pm$ 0.1   & \textit{88.8 $\pm$ 0.2}  &\textbf{ 89.1 $\pm$ 0.1}  \\ \hline
\multicolumn{7}{|c|}{gender-6}                                                                                                                                                                         \\ \hline
Err & 15.4                                                                & 7.97 & 6.07 $\pm$ 0.02                                                      & 6.05 $\pm$ 0.02 & \textit{5.90 $\pm$ 0.02} & \textbf{5.86 $\pm$ 0.02} \\ \hline
\multicolumn{7}{|c|}{gender-22}                                                                                                                                                                        \\ \hline
Err & 10.1                                                                & 5.15 & 3.85 $\pm$ 0.03                                                      & 3.83 $\pm$ 0.03 & \textit{3.70 $\pm$ 0.02} & \textbf{3.66 $\pm$ 0.01} \\ \hline
\end{tabular}
\end{table}

\subsection{Results}

\subsubsection{Accuracy comparisons}
The test error and AUC comparisons are reported in Table \ref{tab:acc}. 

\noindent\textbf{CIFAR100} Compared to the vanilla Resnet model (Base), additive modality combination (Add) does not necessarily help improve test error. Particularly, it helps Resnet-32 but not Resnet-110. It might be due to overfitting on Resnet-110 as there is already much more parameters.

Multiplicative training (Mul), on the contrary, helps reduce error rates in both models. It demonstrates better capability of extracting signals from different modalities. 

Further, MulMix and MulMix*, which is designed to combine the strengths of additive and multiplicative combination, give significant boost in accuracy on both models.

\noindent\textbf{HIGGS} Either fusion model or additive combination gives significant error rate reduction compared to single modality. This is expected as it is very intuitive to aggregate low and high level feature modalities.

Compared to Add, Multiplicative combination has clearly better results on higgs-small but slightly worse results on higgs-full. This can be explained by the fact models are more prone to overfit on smaller datasets and multiplicative training does reduce the overfit. 

Finally, MulMix and MulMix* give significant boost on both small and full datasets. 

\noindent\textbf{Gender} It gives the most dramatic improvements to combine multiple modalities here due to the high level noise in each modalities. Add achieves less than half error rates than what the best single modality could achieve. As a comparison, Mul has similar (slightly better) results. This suggests the two methods might work with similar mechanisms. However, MulMix and MulMix* clearly outperform Add and Mul, showing the benefits of combining two types of combination strategies.

\subsubsection{Compared to deeper fusion networks}
As the new approach, especially MulMix (or MulMix*), introduces additional parameters in fusion networks in each mixture, one natural question is whether the improvements simply come from the increased number of parameters. We answer the question by running additional experiments with additive combination (Add) models with deeper fusion networks. 

\begin{figure}[t]
\centering
\subfigure[CIFAR100]{
\includegraphics[width=.3\textwidth]{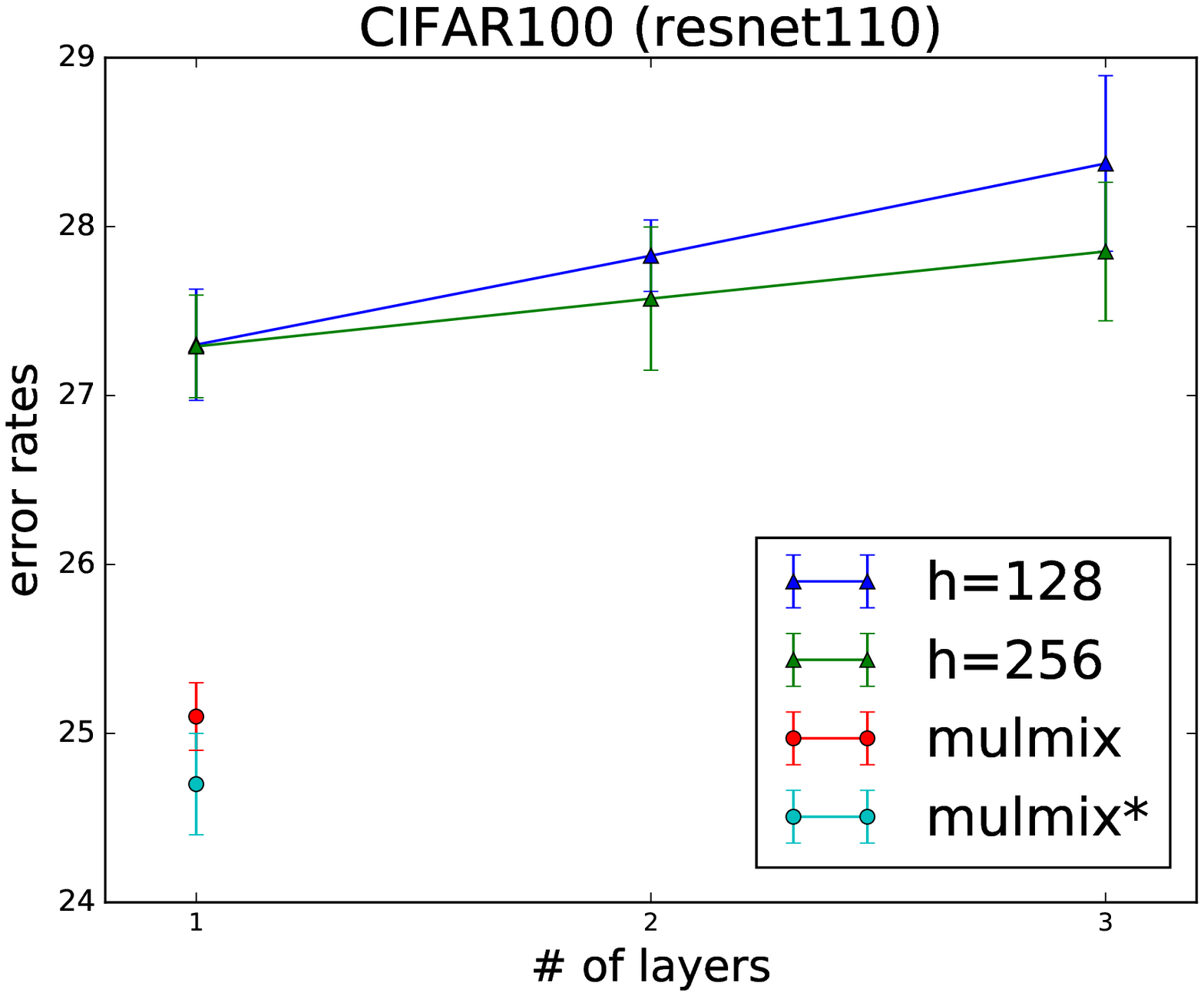}
}
\hfill
\subfigure[HIGGS]{
\includegraphics[width=.3\textwidth]{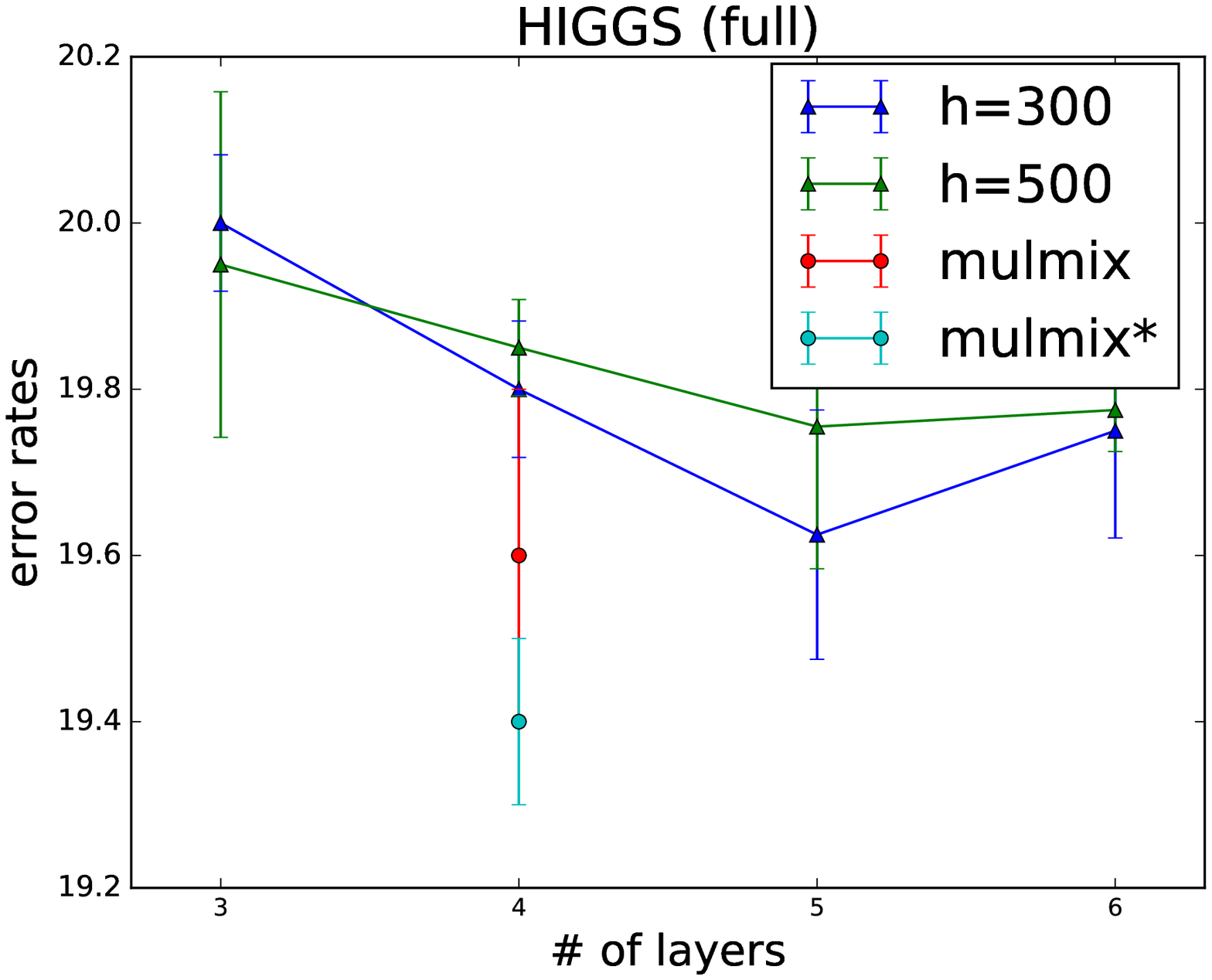}
}
\hfill
\subfigure[gender]{
\includegraphics[width=.3\textwidth]{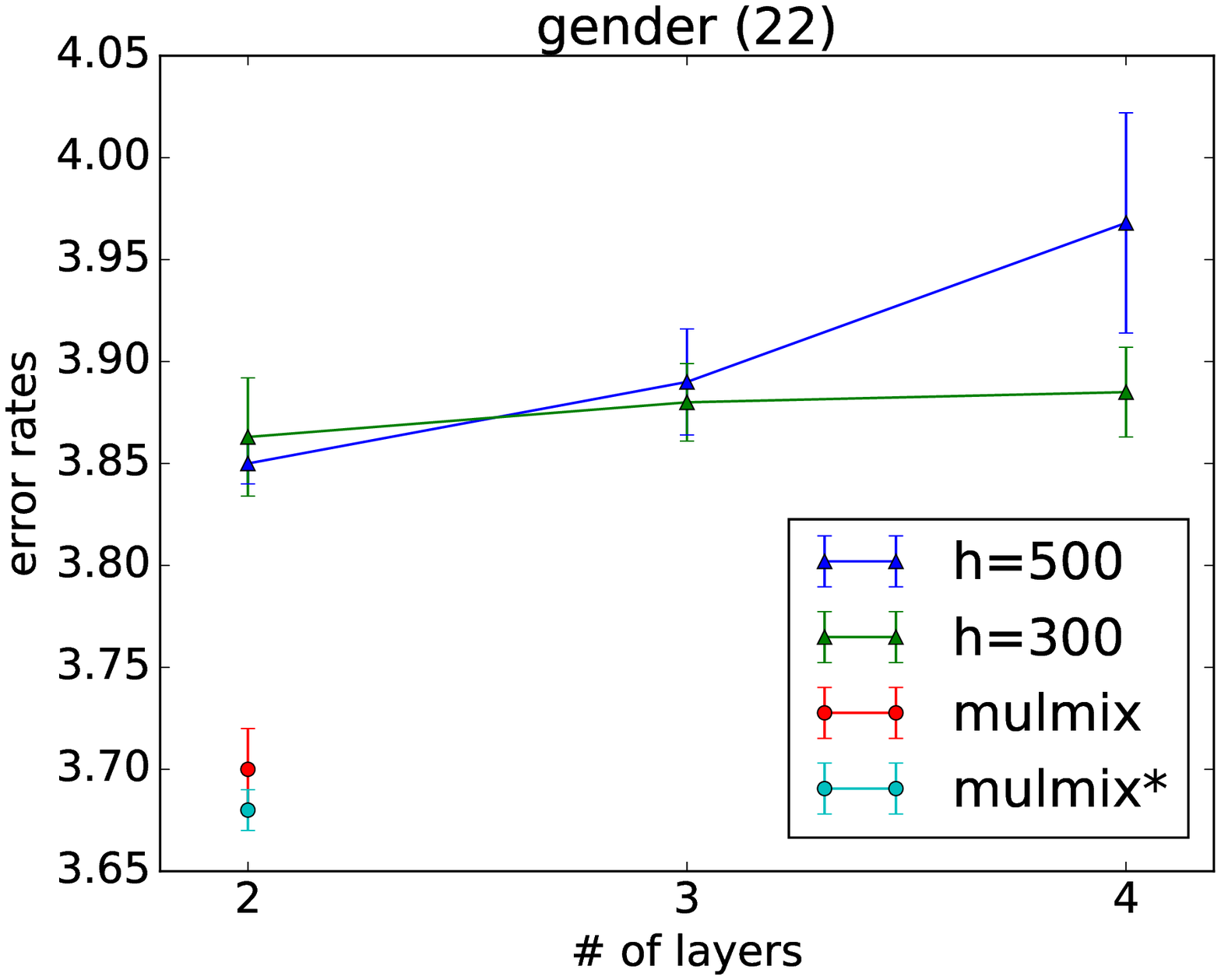}
}

\caption{Comparisons to results from deeper networks. Error rates and standard deviations from fusion networks with hidden layer structures are reported and compared to our models (i.e., \AddMul, and \AddMul*). Simply going deep in networks does not necessarily help improve generalization. Experimental results are from 5 random runs.}
\label{fig:deep}
\end{figure}

The results are plotted in Figure \ref{fig:deep}. We see on CIFAR100 and gender, networks with increased depth lead to worse results. The can be either due to increased optimization difficulty or overfitting. On HIGGS, increased depth first leads to slight improvements and then the error rates go up again. We see even the results at the optimal network depth are not as good as our approach. Overall the figures show that it is the design rather than the depth of the fusion networks that holds their performance. On the contrary our approaches are explicitly designed to extract signals selectively and collectively across modalities.

\subsubsection{Multiplicative combination or model averaging}
\begin{figure}[t]
\centering
\subfigure[CIFAR100]{
\includegraphics[width=.3\textwidth]{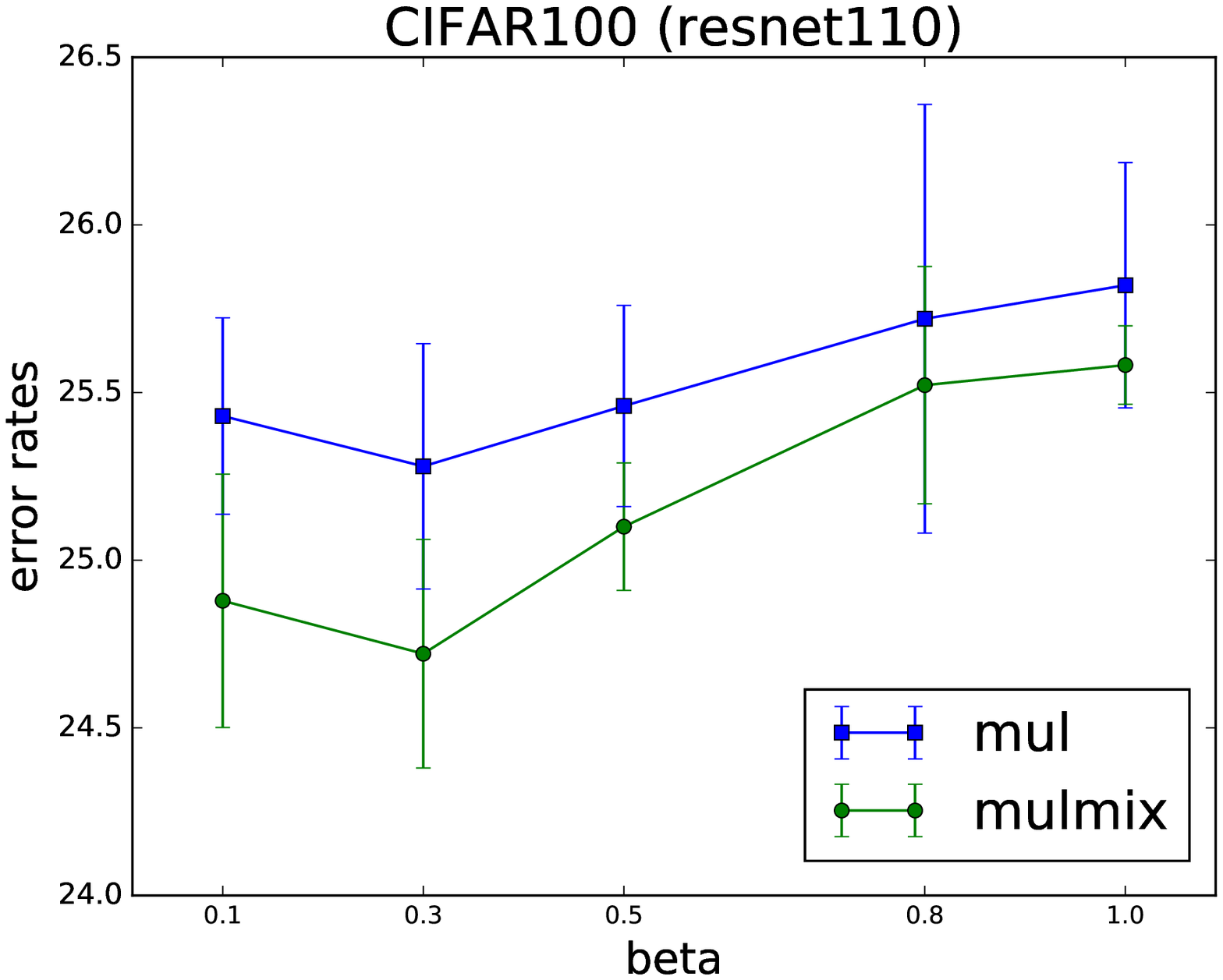}
}
\hfill
\subfigure[HIGGS]{
\includegraphics[width=.3\textwidth]{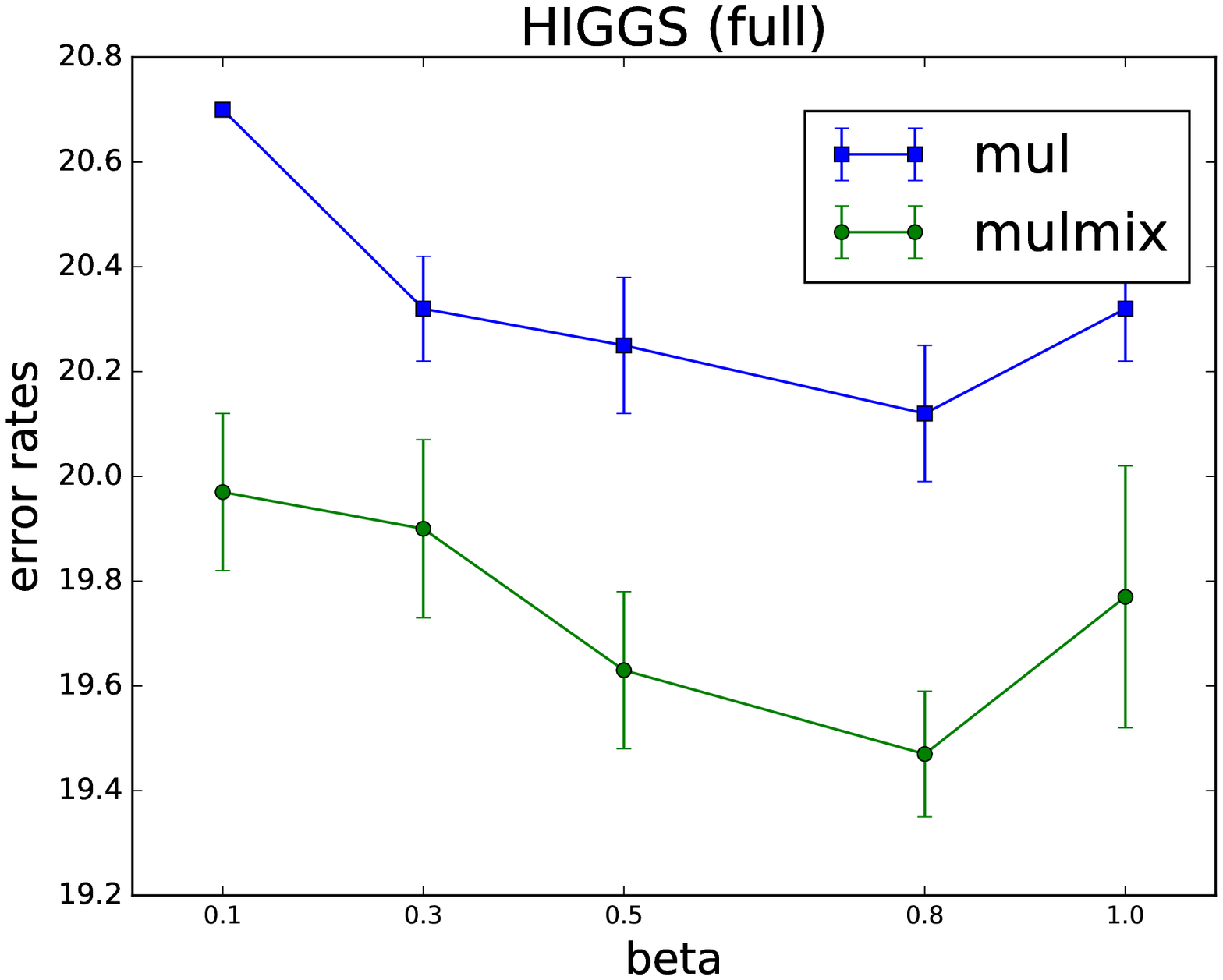}
}
\hfill
\subfigure[gender]{
\includegraphics[width=.3\textwidth]{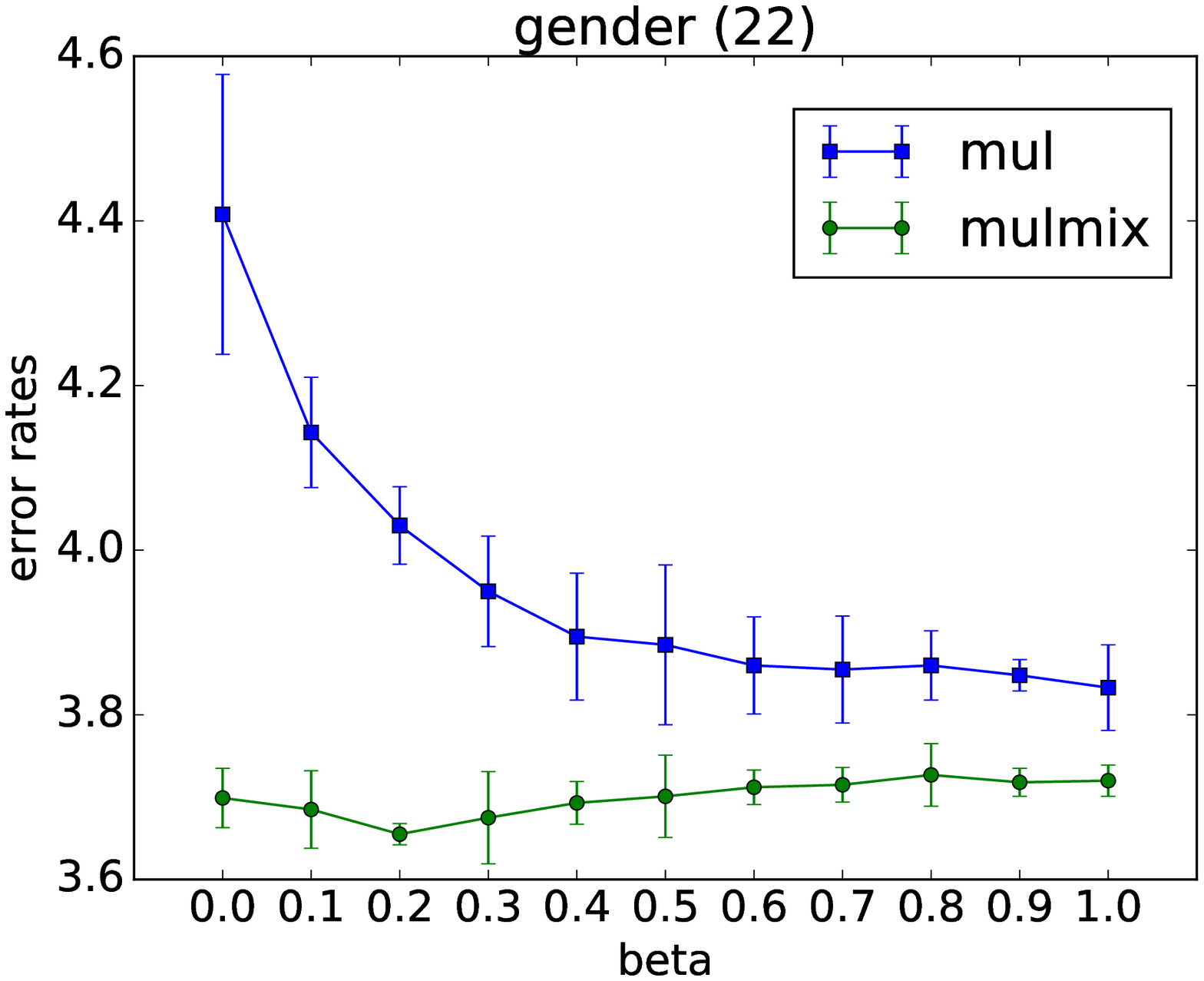}
}

\caption{Error rates and standard deviations under different $\beta$ values. Optimal results do not appear at either 0 or 1. Experimental results are from 5 random runs.}
\label{fig:beta}
\end{figure}

Our loss function in (\ref{eq:scale}) implements a trade-off between model averaging and multiplicative combination. $\beta$=0 makes it an model averaging of different modalities (or modality mixtures) while $\beta$=1 makes the model a non-smoothed multiplicative combination.  To understand the exact working mechanism and to achieve the best results, we tune $\beta$ between 0 and 1 and plot corresponding error rates on different tasks in Figure \ref{fig:beta}.

We observe that the optimal results do not appear at either end of the spectrum. On the contrary, smoothed multiplicative combinations with optimal $\beta$s achieve significantly better results than pure ensemble or multiplicative combination. On CIFAR100 and HIGGS, we see optimal $\beta$ values 0.3 and 0.8 respectively and they are consistent across models Mul and MulMix. On gender, Mul clearly favors $\beta$ to be 1 as each single modality is very noisy and it makes less sense to evenly average predictions from different modalities. 

We do not have a clear theory how to choose $\beta$ automatically. Our hypothesis is that smaller $\beta$ leads to stronger regularization (due to smoothed scaling factors) while larger $\beta$ gives more flexibility in modeling (highly non-linear combination). As a result, we recommend choosing smaller $\beta$ when the original models overfit and larger $\beta$ when the models underfit.

\begin{table}[]
\caption{Error rate results of boosted training (\AddMul) on HIGGS-full and gender-22.}
\label{tab:boosted}
\centering
\begin{tabular}{|c|c|c|c|} \hline
             & $\beta$*           & $\beta$=1.0 (vanilla)        & $\beta$=1.0 (boosted) \\ \hline
HIGGS (full) & 19.5 $\pm$  0.1 & 19.8 $\pm$ 0.3  & 19.5 $\pm$ 0.1     \\ \hline
gender-22    & 3.66 $\pm$ 0.01 & 3.72 $\pm$ 0.02 & 3.67 $\pm$ 0.01   \\ \hline
\end{tabular}
\end{table}
\subsubsection{Boosted training}
We also validate the effectiveness of boosted training technique. We find that when $\beta$ in (\ref{eq:scale}) is not tuned, boosted training significantly improves the results. Table \ref{tab:boosted} shows MulMix($\beta=1$) test errors on HIGGS and gender. Boosted training helps MulMix($\beta=1.0$) achieve almost identical results as MulMix*. It is interesting to see the second and fourth columns have very close numbers. We conjecture the smoothing effect of $\beta$ makes the ``mismatch' issue discussed in Section \ref{sec:boosted} less severe.

\subsubsection{Additional Experiments}

\noindent\textbf{Where the improvements are made?} We are interested in seeing where the improvements are made on this prediction task. It is known that ensemble-like methods help correct predictions on examples where individual classifiers make wrong predictions. However, they also make mistakes on the examples where individual classifiers are correct. This is in general due to overfitting and we call it "over learning."

We expect our methods to reduce errors of "over learning" due to its regularization mechanism -- we tolerate incorrect predictions from a weak modality while preserving its correct predictions. We analyze the errors only on the examples \textit{where individual modalities could make correct predictions}. More clearly, we evaluate the errors on the examples which at least one single modality predicts correctly. 

The result is reported in Table \ref{tab:analysis_gender}. We see~\Mul~and~\AddMul*~both make less ``over learning'' mistakes. Interestingly, the improvement of \AddMul here (0.18) is very close to the improvement on the entire dataset. It suggests our new methods do prevent individual modalities from "over learning."

\begin{table}[]
\centering
\caption{Gender-22 error analysis: Mistakes that multimodal methods make where individual modals do not (we call ``over-learn''); Mul and MulMix* improve Add on that. The overall improvement is very close to the improvement from ``over-learn''.}
\label{tab:analysis_gender}
\begin{tabular}{|c|c|c|c|c|} \hline
  & \Add & \Mul & \AddMul* & improve\\ \hline
 overall    & 3.85 & 3.83 & 3.66 & 0.19\\ \hline
 Over-learn & 2.90 & 2.87 & 2.72 & 0.18\\ \hline
\end{tabular}
\end{table}

\noindent\textbf{Compared to attention models} We also tried attention methods \cite{moon2018multimodal} where attention modules are applied to each modality before additively combined. We experimented on gender prediction because on this task it is most common to see missing modalities. The results are reported in Table \ref{tab:attend}. We do not observe clear improvements.

\begin{table}[]
\centering
\caption{Test errors of attention models on gender tasks.}
\label{tab:attend}
\begin{tabular}{|c|c|c|} \hline
          & Add & Add-Attend \\ \hline
gender-6  &  6.07 $\pm$ 0.02   &   6.07 $\pm$ 0.01     \\ \hline
gender-22 &  3.85 $\pm$ 0.03   &   3.86 $\pm$ 0.03 \\ \hline
\end{tabular}
\end{table}

\noindent\textbf{Compared to CLDL\cite{jin2016collaborative} on CIFAR100} Specific to image recognition domain, CLDL\cite{jin2016collaborative} is one specialization of our ``Mul'' approach based on NIN~\cite{lin2013network}. We implemented CLDL on Resnet. The error rates on two models are 29.6 $\pm$ 0.5 and 25.8 $\pm$ 0.4, respectively.

\section{Conclusion}

This paper investigates new ways to combine multimodal data that accounts for heterogeneity of modality signal strength across modalities, both in general and at a per-sample level. We focus on addressing the challenge of ``weak modalities'': some modalities may provide better predictors on average, but worse for a given instance. To exploit these facts, we propose multiplicative combination techniques to tolerate errors from the weak modalities, and help combat overfitting. We further propose multiplicative combination of modality mixtures to combine the strength of proposed multiplicative combination and existing additive combination. Our experiments on three different domains demonstrate consistent accuracy improvements over the state-of-the-art in those domains, thereby demonstrating the fact that our new framework represents a general advance that is not limited to a specific domain or problem.

\bibliographystyle{abbrv}
\bibliography{ref}

\end{document}